\pdfoutput=1

\documentclass[11pt]{article}

\usepackage[final]{acl}

\usepackage{times}
\usepackage{latexsym}

\usepackage{amsmath}
\usepackage{siunitx}
\usepackage{booktabs}   
\usepackage{multirow}   
\usepackage{siunitx}    
\usepackage{graphicx}   
\usepackage{caption}    
\usepackage[normalem]{ulem} 
\usepackage{enumitem}

\usepackage[T1]{fontenc}

\usepackage[utf8]{inputenc}

\usepackage{microtype}
\usepackage{amsmath}
\usepackage{graphicx}
\usepackage{caption}    
\usepackage{subcaption} 
\usepackage{listings}
\usepackage{xcolor}
\usepackage{tcolorbox}
\usepackage[utf8]{inputenc}
\usepackage{relsize}

\usepackage{graphicx}
\usepackage{listings}
\title{ConstraintLLM: A Neuro-Symbolic Framework for Industrial-Level Constraint Programming}

\author{
 \textbf{Weichun Shi\textsuperscript{1,6}\thanks{\ Equal contribution.}},
 \textbf{Minghao Liu\textsuperscript{2}\footnotemark[1]},
 \textbf{Wanting Zhang\textsuperscript{3}},
 \textbf{Langchen Shi\textsuperscript{4,6}}
 \\
 \textbf{Fuqi Jia\textsuperscript{4,6}},
 \textbf{Feifei Ma\textsuperscript{1,5,6}\thanks{\ Corresponding authors. \texttt{\{maff,zj\}@ios.ac.cn}}},
 \textbf{Jian Zhang\textsuperscript{1,4,6}\footnotemark[2]}
\\
 \textsuperscript{1}Hangzhou Institute for Advanced Study, UCAS, Hangzhou, China
\\
 \textsuperscript{2}University of Oxford, Oxford, UK
\\
 \textsuperscript{3}University of Science and Technology Beijing, Beijing, China
\\
 \textsuperscript{4}SKLCS and Key Laboratory of System Software, ISCAS, Beijing, China
\\
 \textsuperscript{5}Laboratory of Parallel Software and Computational Science, ISCAS, Beijing, China
\\
 \textsuperscript{6}University of Chinese Academy of Sciences, Beijing, China 
\\
   \texttt{shiweichun24@mails.ucas.ac.cn, minghao.liu@cs.ox.ac.uk}
}

\begin{document}
\maketitle

\begin{abstract}
Constraint programming (CP) is a crucial technology for solving real-world constraint optimization problems (COPs), with the advantages of rich modeling semantics and high solving efficiency.
Using large language models (LLMs) to generate formal modeling automatically for COPs is becoming a promising approach, which aims to build trustworthy neuro-symbolic AI with the help of symbolic solvers.
However, CP has received less attention compared to works based on operations research (OR) models.
We introduce ConstraintLLM, the first LLM specifically designed for CP modeling, which is trained on an open-source LLM with multi-instruction supervised fine-tuning.
We propose the Constraint-Aware Retrieval Module (CARM) to increase the in-context learning capabilities, which is integrated in a Tree-of-Thoughts (ToT) framework with guided self-correction mechanism.
Moreover, we construct and release IndusCP, the first industrial-level benchmark for CP modeling, which contains 140 challenging tasks from various domains.
Our experiments demonstrate that ConstraintLLM achieves state-of-the-art solving accuracy across multiple benchmarks and outperforms the baselines by 2x on the new IndusCP benchmark. Code and data are available at: \url{https://github.com/william4s/ConstraintLLM}.

\end{abstract}

\section{Introduction}
\label{sec:intro}

Constraint Optimization Problems (COPs) are prevalent in real-world applications, such as scheduling, resource allocation, routing, and logistics optimization \cite{puget1995applications,wallace1996practical,simonis1999building}.
\textbf{Constraint Programming (CP)} is a powerful paradigm for solving these complex combinatorial problems.
Traditionally, CP involves two main steps: modeling and solving \cite{marriott1998programming}.

The goal of the modeling step is to translate a real-world problem into a formal constraint satisfaction or optimization problem, which is typically defined by a set of variables, domains for these variables, and constraints that must be satisfied among them (Please refer to Appendix~\ref{sec:appendix_background} for a broader background on symbolic reasoning paradigms).

The significance and uniqueness of CP are underscored by its fundamental differences from OR and mainstream Mathematical Optimization (MO) approaches, such as Linear Programming and Integer Programming. Unlike OR/MO, which primarily relies on algebraic formulations, CP fundamentally employs a \textbf{declarative approach}. This means developers focus more on clearly \textit{describing} the problem's inherent structure and constraints, rather than pre-specifying the concrete steps to find a solution. CP's strong expressive power allows for the natural and compact modeling of complex combinatorial structures and logical conditions, A representative example is presented in Figure ~\ref{fig:circular_permutation_problem} (in Appendix). This is particularly effective for problems such as scheduling and resource allocation, which can become exceedingly cumbersome or result in overly large models if expressed purely using LP/IP formulations, leading to less intuitive and potentially more complex implementations. This declarative nature, establishes CP as an indispensable and highly valuable alternative optimization paradigm for tackling industrial challenges.

However, despite CP's powerful modeling capabilities, the manual CP modeling process can be time-consuming, error-prone, and requires significant domain expertise \cite{freuder2014grand,o2010automated}.
Recently, to overcome these limitations, researchers have begun to explore the use of Large Language Models (LLMs) for automated or semi-automated constraint modeling. 
For instance, CP-LLM-ICL \cite{michailidis2024constraint} exemplifies this trend; however, this work relies solely on RAG for CP modeling without leveraging Supervised Fine-Tuning (SFT) to improve the model's inherent modeling abilities.

Despite some progress in LLM-based constraint modeling, existing approaches still face several challenges when dealing with tasks of industrial-level complexity.
These challenges include the potential for generated models to contain syntactic errors, logical inconsistencies, or fail to effectively capture the core constraints of the problem description.
Furthermore, current methods may struggle to generalize to diverse and large-scale industrial application scenarios.

\citet{hao2024large} also explore using LLMs for formal modeling, but our approach differs significantly from theirs. Their method relies on an interactive, prompt-driven workflow using static examples for the specific domain of travel planning. In contrast, we introduce a fully automated pipeline powered by a fine-tuned model and a dynamic retrieval module (CARM). Furthermore, we validate our approach on the broad, multi-domain IndusCP benchmark, addressing a much wider scope of industrial problems.

To address these challenges, we propose ConstraintLLM, a powerful LLM-based neuro-symbolic framework specifically designed to establish and solve industrial-level CP models.
We design a Constraint-Aware Retrieval Module (CARM) to replace the embedding-based retrieval component typically found in standard Retrieval-Augmented Generation (RAG) frameworks.
Furthermore, we integrate CARM into Tree-of-Thoughts (ToT) \cite{yao2023tree}.
To address both code and logical errors in the model's output, we employ an Iterative Self-Correction mechanism with Guided Retrieval.
Finally, we perform multi-instruction SFT on an open-source LLM to comprehensively enhance its capabilities throughout the CP model construction process.
Our neuro-symbolic approach deeply integrates the strengths of LLMs in understanding and generation with the strengths of symbolic solvers in precise solution and verification.

To comprehensively evaluate the performance of ConstraintLLM, we introduce IndusCP, the first industrial-level benchmark for CP modeling.
This benchmark comprises 140 curated problem instances from various domains, including scheduling, packing, and pathfinding, designed to reflect the complex scenarios encountered in industrial applications.

Our primary contributions can be summarized as follows: 





\begin{enumerate}[label=(\arabic*)]
    \item To the best of our knowledge, we are pioneering the training of open-source LLMs specifically for CP modeling (Section~\ref{subsec:multi_instruction_sft}).
    \item We construct and release the IndusCP benchmark (Section~\ref{sec:indus_cp_benchmark}), the first benchmark of its kind designed to assess LLMs on solving real-world, industrial-level CP problems.
    \item We innovatively propose CARM (Section~\ref{subsec:Constraint-Aware Retrieval Module}), which significantly enhancing the In-Context Learning (ICL) capabilities of LLMs by providing contextual exemplars that are better aligned with the target problem in terms of logical structure and mathematical principles.
    \item Our ConstraintLLM demonstrates state-of-the-art performance (Section~\ref{sec:SOTA}) across multiple COP benchmarks, including NL4OPT, LGPs, LogicDeduction, and the more challenging IndusCP.
\end{enumerate}

\section{Methodology}
\label{sec:method}

\begin{figure*}
    \centering 
    \includegraphics[width=2\columnwidth]{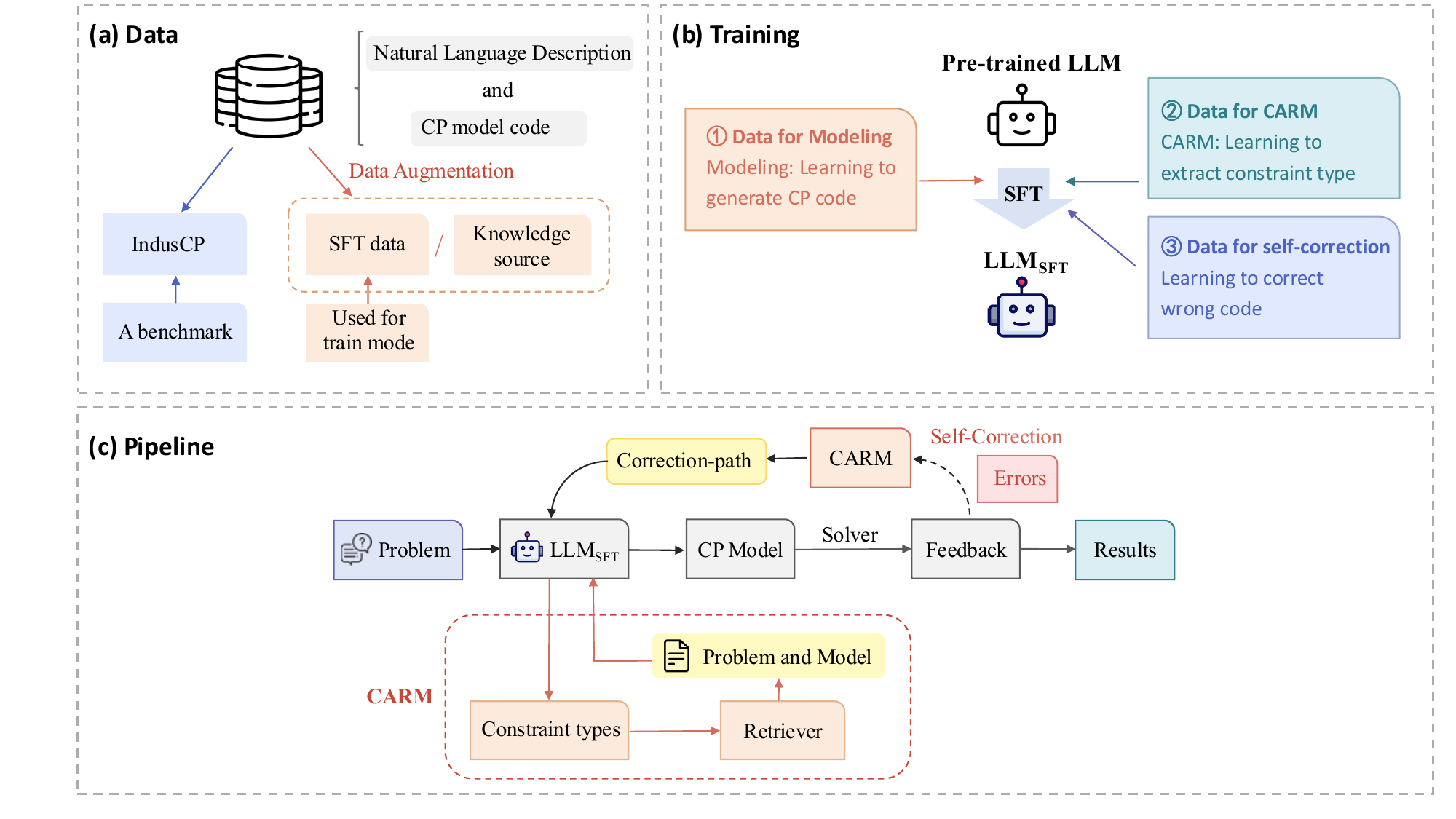} 
    \caption{The framework of ConstraintLLM.
(a) Illustrates the process of constructing IndusCP and preparing the training data.
(b) Depicts the training phase, where we employ multi-instruction SFT to teach the model skills in CP modeling, constraint type extraction, and self-correction.
(c) Outlines the inference pipeline of ConstraintLLM: a problem, augmented with ICL examples provided by CARM, is fed to the LLM to formulate a CP model. This CP model is then processed by the solver. The feedback from the solver is used to either derive the final answer or to initiate a self-correction process.}
    \label{fig:framework} 
\end{figure*}

To enhance the capability of LLMs in modeling and solving Constraint Satisfaction/Optimization Problem (CSP/COP), we propose a comprehensive framework centered on Multi-Instruction Supervised Fine-Tuning (SFT, Section~\ref{subsec:multi_instruction_sft}) to bolster model performance on key sub-tasks: constraint extraction, model generation, and self-correction. Diverging from traditional RAG, we introduce a \textbf{Constraint-Aware Retrieval Module (CARM, Section~\ref{subsec:Constraint-Aware Retrieval Module})}. CARM analyzes the constraint profile of a problem to retrieve exemplars with stronger logical and mathematical relevance in problem structure and solving logic, thereby guiding subsequent ICL. This module is further integrated with a Constraint-Aware Exploration with Tree-of-Thoughts (ToT) framework (Section~\ref{subsec:Constraint-Aware_Exploration_with_Tree-of-Thoughts}), enabling the main modeling LLM ($L_{\text{coder}}$) to systematically explore diverse modeling paths, with search optimized by evaluating branch models on test cases.

To address code errors, we design an Iterative Self-Correction with Guided Retrieval mechanism (Section~\ref{subsec:Iterative Self-Correction with Guided Retrieval}), activated upon external solver validation failure. This mechanism leverages CARM to retrieve relevant exemplars from a case library $\mathcal{E}$ containing correction paths, guiding the LLM in repairing the code. The entire framework undergoes multi-instruction SFT using Parameter-Efficient Fine-Tuning (PEFT) techniques, utilizing meticulously constructed data for each sub-task to comprehensively enhance the model's proficiency in solving constraint problems.
Overall, our framework integrates the powerful generative and understanding capabilities of LLMs with the precise validation of external symbolic solvers and the explicit utilization of structured knowledge , thereby constituting a neuro-symbolic approach aimed at more robustly and efficiently solving complex problems.

To provide a clearer illustration of our pipeline, we present a step-by-step walkthrough of solving the Traveling Salesman Problem (TSP) in Appendix~\ref{sec:appendix_walkthrough}.

\subsection{Constraint-Aware Retrieval Module}
\label{subsec:Constraint-Aware Retrieval Module}
\paragraph{Motivation.}
Retrieval-augmented generation (RAG) \cite{lewis2020retrieval} is a successful technique for boosting LLM performance on complex, knowledge-intensive tasks. However, conventional exemplar retrieval methods, whether based on keyword matching or dense vector similarity, primarily focus on lexical overlap or generalized semantic proximity. 
This limitation is particularly acute for CP modeling; indeed, our experiments showed that standard prompting methods, including Chain-of-Thought and conventional RAG, struggle to handle the logical complexity of industrial-scale problems (see results in Table~\ref{tab:llm_comprehensive_comparison}). 

To bridge this gap, we designed the Constraint-Aware Retrieval Module (CARM) to operate on a deeper, more structural level. Instead of matching surface-level text, CARM is designed to capture the underlying \textbf{logical structure} of problems. It achieves robust generalization by matching problems based on their "constraint profiles"—the core set of constraints required for a valid model. For example, a factory scheduling problem and a nurse rostering problem, despite their vastly different textual descriptions, might both rely on \texttt{Cumulative} and \texttt{AllDifferent} constraints. By retrieving exemplars based on this shared logical foundation, CARM can generalize effectively across diverse domains, providing the model with truly relevant reasoning patterns.

\paragraph{Module Design.}
To effectively enhance the ICL capabilities of the primary modeling LLM (denoted as $L_{\text{modeling}}$, designated for generating the solution, e.g., constraint modeling code or planning steps) when tackling complex problems, we design and integrate a \textbf{CARM}.
The core objective of this module is to transcend traditional representation similarity-based retrieval paradigms.
By performing a deeper analysis of the problem's intrinsic logical structure, specifically its \textbf{Constraint Patterns}, it aims to accurately match and provide $L_{\text{modeling}}$ with exemplars that exhibit high similarity at the constraint level.
We posit that the constraints form the essential backbone of a problem's solution logic; thus, retrieving exemplars based on constraint pattern similarity is more likely to ensure their effectiveness and relevance.
The module comprises the following two key steps:

\paragraph{Step 1. LLM-Powered Constraint Type Extraction}

This step aims to transform the input natural language problem description $Q_{\text{NL}}$ into structured constraint information, which we denote as $C$. We employ an auxiliary LLM, $L_{\text{analyzer}}$ (which could be a general-purpose LLM guided by specific prompts $P$, or a fine-tuned specialized model), acting as a \textbf{Semantic Parser}. This parsing process maps $Q_{\text{NL}}$ to its structured representation $C$.
This transformation can be formalized as:
\begin{equation} \label{eq:semantic_parsing}
C ={L_{\text{analyzer}}}(Q_{\text{NL}}, P)
\end{equation}
where the output $C(Q)$ is a set of constraint types drawn from a predefined constraint ontology $\mathcal{O}$:
\begin{equation}
C(Q_\text{NL}) = \{c_1, c_2, ..., c_n\} \quad \forall i \in \{1, ..., n\}, c_i \in \mathcal{O} 
\end{equation}

Here, $Q_{\text{NL}}$ denotes the input query problem described in natural language. The goal is to identify the underlying constraint types within the query, collectively represented as $C(Q)$, the Constraint Profile of $Q$. Specifically, $C(Q_{\text{NL}}) = \{c_1, c_2, ..., c_n\}$, where each $c_i$ denotes the $i$-th identified constraint type, and $n$ is the total number of such types, with $n \ge 0$. These constraint types are drawn from a predefined Constraint Ontology $\mathcal{O}$, which includes categories such as \texttt{AllDifferent}, \texttt{Cumulative}, \texttt{LexDecreasing}, \texttt{NoOverlap}, and others.

To extract these constraints, the model $L_{\text{analyzer}}$ is guided by carefully designed prompts to perform deep semantic analysis on $Q_{\text{NL}}$. Through this process, various constraints are identified and mapped to standardized types within $\mathcal{O}$, resulting in the final Constraint Profile $C(Q)$.

\paragraph{Step 2. Constraint Profile-Driven Similarity Matching and Retrieval}

In this step, the extracted constraint profile $C(Q)$ is used to identify the most relevant cases from a pre-built case library $\mathcal{D} = \{D_1, D_2, \dots, D_m\}$. Each case $D_j$ in the library consists of its original description $D_{j,\text{NL}}$ and the corresponding reference solution $D_{j,\text{Sol}}$, with its own pre-computed and indexed constraint profile $C(D_j)$.

To determine which cases are most relevant, the similarity between the query’s constraint profile $C(Q)$ and each case’s constraint profile $C(D_j)$ is evaluated using the Jaccard similarity coefficient. This coefficient measures the overlap between the sets of constraint types:
\begin{equation}
\text{Sim}(C(Q), C(D_j)) = \frac{|C(Q) \cap C(D_j)|}{|C(Q) \cup C(D_j)|}
\end{equation}

Once the similarity scores are computed, the cases in $\mathcal{D}$ are ranked in descending order. The top-$k$ cases, $\{D_{r_1}, D_{r_2}, \dots, D_{r_k}\}$, are then selected as the final set of relevant exemplars to be provided to $L_{\text{modeling}}$.

By employing CARM, we aim to provide the main model $L_{\text{modeling}}$ with exemplars that are not only potentially relevant in terms of surface text but, more crucially, possess stronger logical and mathematical relevance in terms of problem structure and solving logic,as reflected by the constraints. This strategy, based on deep logical matching, aims to maximize the efficiency of ICL, guiding $L_{\text{modeling}}$ towards more effective Analogical Reasoning and solution strategy transfer, thereby improving its performance on the target complex tasks.

\subsection{Constraint-Aware Exploration with Tree-of-Thoughts}
\label{subsec:Constraint-Aware_Exploration_with_Tree-of-Thoughts}
\paragraph{Motivation.}
Constraint modeling code is characterized by its highly structured nature, strict syntax, and strong logical requirements. Its generation process is essentially a complex search problem involving numerous decision points. To systematically explore this vast search space, we introduce a Tree-of-Thoughts (ToT) framework and integrate it with our proposed \textbf{CARM}. This module replaces traditional embedding-based retrieval, enabling more effective guidance of the exploration process.

\paragraph{Exploring Diverse Modeling Choices.}
The ToT framework simulates modeling via an exploration tree, rooted at the initial problem or an empty model. From any node representing a partial solution, ToT generates parallel "thoughts." In our context, a "thought" is a concrete modeling decision or code snippet extending the current solution, primarily including:
(1): Global Constraint Selection: Leveraging CARM to explore different global constraints (e.g., \texttt{AllDifferent}, \texttt{Cumulative});
(2): Variable Definition Strategies: Exploring diverse variable definitions (e.g., array vs. named variables, integer vs. Boolean variables);
(3): Auxiliary Variable Introduction: Investigating auxiliary variables to simplify constraints or aid solver pruning, with CARM recommending relevant construction patterns.
CARM is central to this ToT process. Instead of relying on fuzzy semantic similarity, it retrieves syntactically correct, logically relevant code snippets or modeling patterns satisfying specific constraint structures, based on the current modeling context and structured queries. This enables ToT to generate high-quality, diverse "thought branches.
More details of ToT are provided in Appendix \ref{app:detail_of_tot}.

\subsection{Iterative Self-Correction with Guided Retrieval}
\label{subsec:Iterative Self-Correction with Guided Retrieval}

To rectify logical flaws beyond syntactic errors, we employ an iterative self-correction mechanism triggered by failures during validation with external solvers (e.g., Choco, ACE). Solver feedback pinpoints logical inconsistencies, guiding the correction process.

Central to this mechanism is a two-stage retrieval strategy for selecting relevant correction exemplars from a pre-constructed database $\mathcal{E}$. Each exemplar $e \in \mathcal{E}$ contains a problem description, incorrect code, a correction path, and the correct code. Given the current error context $c_{err}$ (derived from solver feedback and erroneous code), we first identify a candidate set $\mathcal{C}$ based on embedding similarity, and then re-rank these candidates using our CARM based on constraint relevance. The top-ranked exemplar $e^*$ is selected as:
\begin{align}
    \mathcal{C} &= \text{Top-k}_{e \in \mathcal{E}} \left( S_{emb}(c_{err}, e) \right) \label{eq:retrieval_stage1} \\
    e^* &= \underset{e \in \mathcal{C}}{\text{argmax}} \left( S_{carm}(c_{err}, e) \right) \label{eq:retrieval_stage2}
\end{align}
where $S_{emb}$ denotes the embedding similarity score (e.g., cosine similarity between embeddings of $c_{err}$ and error info in $e$), and $S_{carm}$ is the relevance score assigned by the CARM, prioritizing exemplars whose correction paths address constraints pertinent to $c_{err}$.

The retrieved exemplar $e^*$ provides an in-context learning example for the LLM. The LLM receives the original problem description, the current incorrect code, the solver feedback, and the components of $e^*$ (incorrect code, correction path, correct code) to generate a revised model code.

This process iterates: (1) Generate code, (2) Validate with solver, (3) If failure, retrieve exemplar $e^*$ using Eqs. \ref{eq:retrieval_stage1}-\ref{eq:retrieval_stage2}, (4) Generate corrected code using LLM with $e^*$ for in-context learning, (5) Repeat validation. We typically perform 4 rounds of iteration, which substantially improves code correctness by enabling the model to learn from targeted feedback and relevant correction strategies.

\subsection{Multi-Instruction Supervised Fine-Tuning}
\label{subsec:multi_instruction_sft}

We employ PEFT with three primary objectives: (1) enhancing the model's core CP modeling capability; (2) improving its ability to accurately extract constraint types from problem descriptions for CARM; and (3) strengthening its capacity to correct syntactic and logical code errors during Self-Correction (Section~\ref{subsec:Iterative Self-Correction with Guided Retrieval}).

Our training data primarily originates from a reserved subset of our IndusCP benchmark, augmented using techniques like EDA \cite{wei2019eda} and variable renaming \cite{yu2022data} for both problem descriptions and code to enhance diversity and robustness.
We specifically construct paired data for constraint type extraction (for CARM) and an exemplar database $\mathcal{E}$ containing \{problem description, incorrect code, correction path, correct code\} (see Section~\ref{subsec:Iterative Self-Correction with Guided Retrieval}) to bolster error correction capabilities.
For a detailed methodology on data augmentation and construction, please refer to Appendix~\ref{app:detail_of_data_augmentation}.


\begin{table*}[tb]
    \caption{Distribution of the IndusCP dataset}
    \label{tab:distribution}
    \centering
    \resizebox{0.65\textwidth}{!}{
    \begin{tabular}{ccc} 
        \toprule
        \textbf{Category} & \textbf{Number} & \textbf{Percentage} \\ \hline
        Scheduling \& Sequencing & 31 & 23.8\%  \\
        Routing \& Logistics & 12 & 9.2\% \\ 
        Resource Allocation \& Assignment & 23 & 17.7\%\\ 
        Layout, Packing \& Cutting & 10 & 7.7\% \\ 
        Design \& Configuration & 16 & 12.3\% \\ 
        Combinatorial Puzzles \& Games & 21 & 16.2\%\\
        Data-Driven Optimization \& Analytics & 6 & 4.6\%\\
        Cryptography \& Algorithmic Puzzles & 4 & 3.1\%\\
        Manufacturing \& Production Planning & 4 & 3.1\%\\
        Telecommunications \& Network Design  & 2&  1.5\%\\ 
        Others  & 1 & 0.8\%\\ 
        \bottomrule
    \end{tabular}
    }
\end{table*}
\section{IndusCP: A New Industrial-Level Benchmark for Constraint Satisfaction Problems}
\label{sec:indus_cp_benchmark}

Recent advancements have applied LLMs to solve CP problems \cite{michailidis2024constraint}. However, these initial explorations often focus on specific problem types or simplified settings, not yet fully addressing the diverse and generalized scenarios encountered in practical, industrial-scale CP applications. This gap is partly due to the limitations of existing benchmarks in comprehensively evaluating LLM capabilities on complex CSPs/COPs.
For instance, while NL4OPT \cite{ramamonjison2023nl4opt} provides natural language descriptions for linear optimization problems, its inherent focus on the linear programming (LP) modeling paradigm and its associated, relatively simpler constraint types renders it unsuitable for assessing an LLM's ability to handle CP. CP are typically characterized by intricate combinatorial structures and a rich variety of constraints, especially powerful global constraints, which are not central to LP formulations.
Similarly, the LGPs dataset \cite{mitra2015learning}, consisting of logical puzzles, primarily involves basic logical and arithmetic constraints (such as \texttt{AllDifferent}, \texttt{Xor}, \texttt{LogicalAnd}, and comparison operators) when formulated as CSPs. While useful for certain reasoning tasks, LGPs also falls short in providing the comprehensiveness and complexity needed to benchmark LLMs on the intricate industrial-grade constraints found in real-world CSPs.

To address these limitations, we introduce \textbf{IndusCP}, a new industrial-level benchmark for CP, specifically designed to enable a comprehensive evaluation of frameworks for solving CP problems, particularly those leveraging LLMs for modeling and solution, and to better reflect real-world application scenarios.
It aims to bridge the significant gap between the relatively simplified CP instances common in academic research and the multifaceted, complex problems encountered in industry. All problems in the IndusCP benchmark are NP-hard.


\begin{table*}[t]
\centering
\small
\caption{Comparison of IndusCP with existing benchmarks.}
\label{tab:benchmark_comparison}
\begin{tabular}{lrrrl}
\toprule
\textbf{Benchmark} & \textbf{\# Probs} & \textbf{Avg. Const.} & \textbf{Avg. Vars.} & \textbf{Problem Type} \\
\midrule
LogicDeduction & 200 & 5.06 & 6.00 & Positional Reasoning \\
LGPs & 100 & 12.62 & 12.00 & Puzzles \\
NL4OPT & 271 & 4.25 & 2.02 & Linear Programming \\
\midrule
\textbf{IndusCP (Ours)} & \textbf{140} & \textbf{240.14} & \textbf{101.24} & \textbf{Diverse Combinatorial} \\
\bottomrule
\end{tabular}
\end{table*}

We compare IndusCP with several existing benchmarks in Table~\ref{tab:benchmark_comparison}. As the table clearly demonstrates, IndusCP operates on a vastly different scale in terms of both the number and complexity of constraints and variables, providing a much more challenging and realistic testbed for evaluating modern solvers.

IndusCP comprises 140 curated problem instances. Each instance includes 2 to 5 distinct test cases to assess model robustness across different scales and conditions. The problems cover a wide range of classic constraint satisfaction and optimization domains, primarily including: Scheduling and Sequencing, Packing and Cutting, Routing and Pathfinding, Allocation and Assignment, and Combinatorial Design and Puzzles. The distribution is shown in
\tablename~\ref{tab:distribution}.

The detailed curation process for the benchmark is described in Appendix~\ref{sec:appendix_construction_details}.

\newcommand{\bestres}[1]{\textbf{\uline{#1}}}
\begin{table*}[htbp]
\centering
\small 
\caption{Performance comparison across models and methods using SA(\%) metric. Best results per dataset are \textbf{\uline{bolded and underlined}}. ``*'' indicates results from original papers, ``‡'' marks previous SOTA. For CoT (one-shot) and RAG/ConstraintLLM (four-shot) methods. Improvement rates show gains over CoT baselines.}
\label{tab:llm_comprehensive_comparison} 
\begin{tabular}{llcccc}
\toprule
\multirow{2}{*}{LLM} & \multirow{2}{*}{Method} & \multicolumn{4}{c}{Benchmarks} \\
\cmidrule(lr){3-6}
& & IndusCP & NL4OPT & LGPs & LogicDeduction \\
& & {SA (\%)} & {SA (\%)} & {SA (\%)} & {SA (\%)} \\
\midrule
\multicolumn{6}{l}{\textit{Baseline Methods}} \\
\multirow{3}{*}{Qwen2.5-Coder-32B} & Direct Solving & 23.3 & 10.1 & 3.3 & 51.0 \\
& CoT (one-shot) & 20.9 & 85.6 & 2.0 & 83.5 \\
& RAG (four-shot) & 24.3 & 91.7 & 86.0 & 17.2 \\
\midrule
\multicolumn{6}{l}{\textit{Previous Work}} \\
CP-LLM-ICL~\cite{michailidis2024constraint} & & 22.7 & 87.5* & 76.0*‡ & 93.5 \\
LLMOPT*~\cite{jiang2024llmopt} & & 3.3 & 93.0*‡ & 0.0 & 16.0 \\
Logic-LM*~\cite{pan2023logic} & & / & / & / & 87.6*‡ \\
\midrule
\multicolumn{6}{l}{\textit{Model Comparisons}} \\
\multirow{3}{*}{ChatGPT-4o} & CoT & 26.5 & 82.3 & 41.0 & 67.0 \\
& RAG & 33.5 & 91.5 & 87.0 & 97.0 \\
& ConstraintLLM (w/o ToT) & 49.8 & 96.3 & 90.0 & 97.0 \\
\midrule
\multirow{3}{*}{DeepSeek-V3} & CoT & 38.5 & 84.9 & 42.0 & 85.0 \\
& RAG & 51.0 & 96.7 & 87.0 & 99.0 \\
& ConstraintLLM (w/o ToT) & \bestres{57.8} & 96.7 & \bestres{92.0} & \bestres{100.0} \\
\midrule
\multirow{4}{*}{ConstraintLLM-32B} & CoT & 17.5 & 85.2 & 2.0 & 84.5 \\ 
& RAG & 21.8 & 88.6 & 82.0 & 92.0 \\ 
& ConstraintLLM (w/o ToT) & 40.0 & 95.2 & 91.0 & 96.0 \\
& ConstraintLLM (w/ ToT) & 51.3 & \bestres{99.26} & \bestres{92.0} & \bestres{100.0} \\
\bottomrule
\end{tabular}
\end{table*}

\section{Experiments}
\label{sec:exp}
\subsection{Experimental Setup}
\paragraph{Evaluation Benchmarks and Metrics.}
\label{sec:dataset&metrics}

We use NL4OPT~\cite{ramamonjison2023nl4opt}, LGPs \cite{mitra2015learning}, LogicDeduction \cite{pan2023logic} and IndusCP as evaluation benchmarks.
IndusCP includes 140 COPs in its test set.
NL4OPT is the most widely used benchmark for linear optimisation. We select 271 instances for the test set and 731 instances for knowledge sources.
LGPs consist of logical puzzles described with clues and entities, which can be formulated as CSPs. We select 50 instances as knowledge sources and 100 instances as test data.
LogicDeduction is a complex logical inference task that can be expressed as CSPs. For our experiments, we randomly select 200 instances for the test set and 100 instances for the knowledge sources.
Detailed descriptions of these datasets are provided in Appendix \ref{app:datasets_detail}.
We use Solving Accuracy (SA) to measure the proportion of problems for which an approach, after its full processing, outputs a verified correct solution.

\begin{figure*}[ht!] 
    \centering 

    \begin{subfigure}[b]{0.48\textwidth} 
        \centering
        \includegraphics[width=\linewidth]{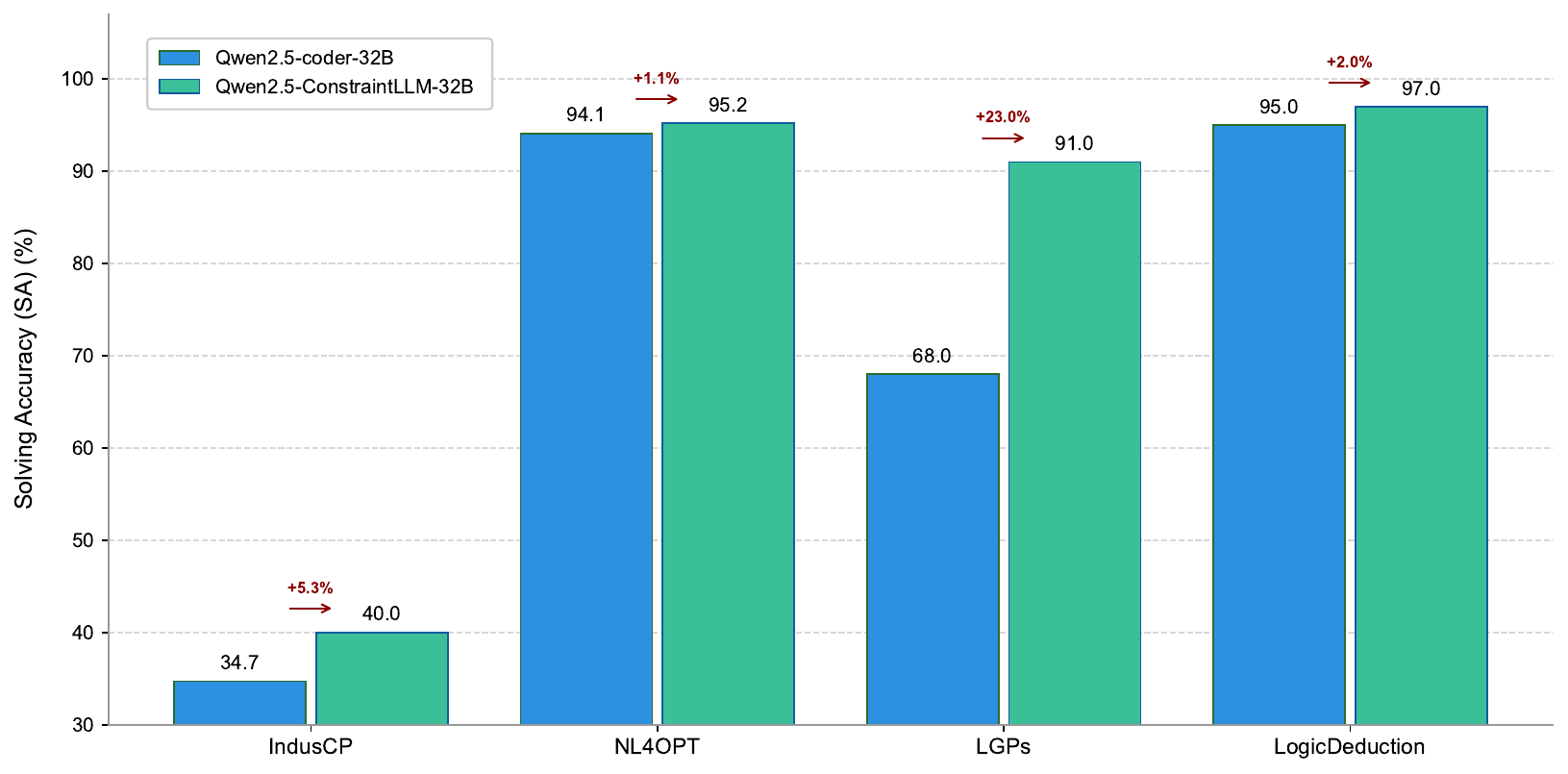}
        \caption{Comparison between Qwen2.5-coder-32B and the model with SFT under ConstraintLLM framework.}
        \label{fig:sub_image1}
    \end{subfigure}
    \hfill
    \begin{subfigure}[b]{0.48\textwidth}
        \centering
        \includegraphics[width=\linewidth]{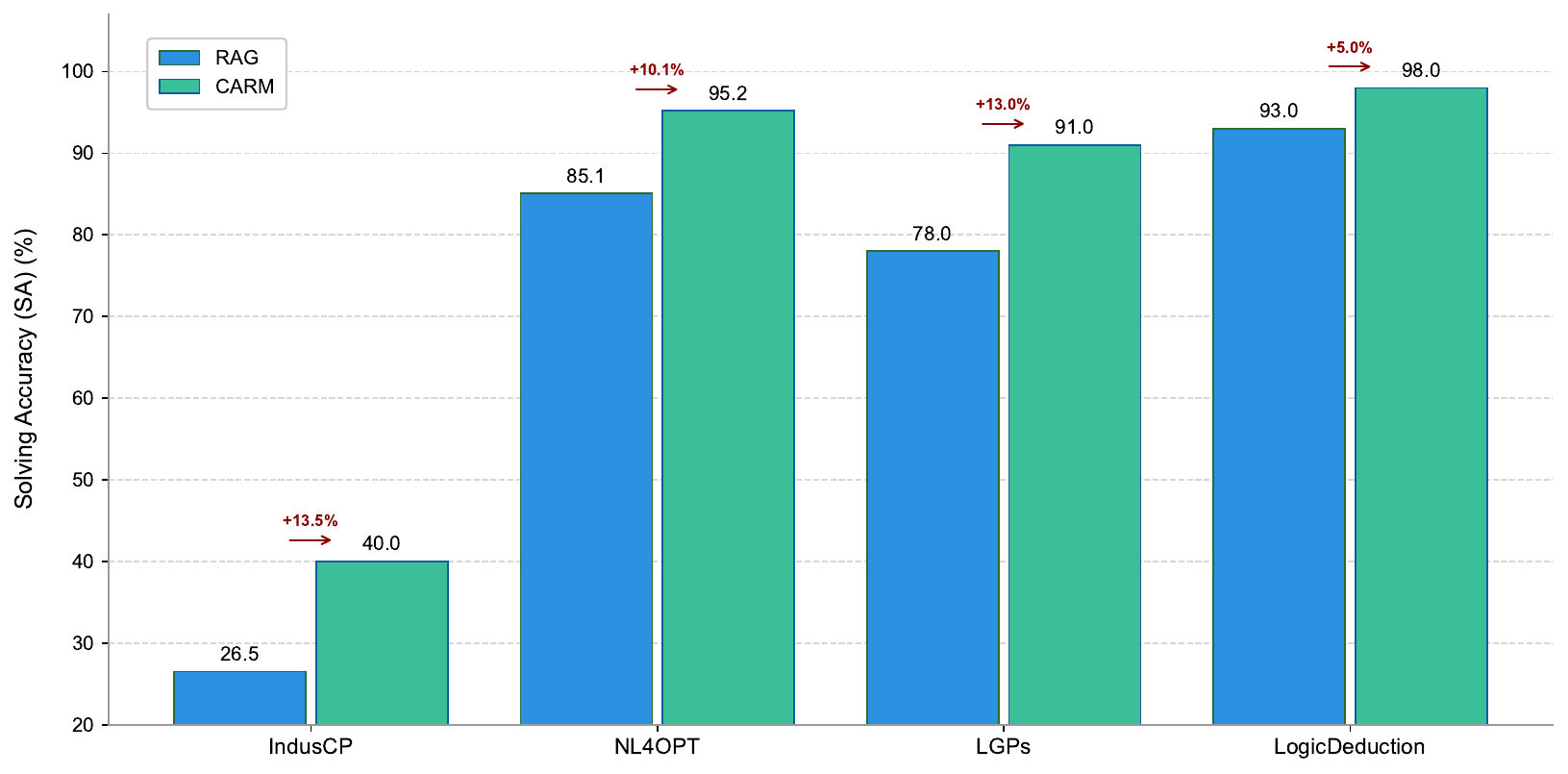}
        \caption{Comparison between RAG and CARM under ConstraintLLM framework.}
        \label{fig:sub_image2}
    \end{subfigure}

    \vspace{0.5cm} 

    \begin{subfigure}[b]{0.48\textwidth} 
        \centering
        \includegraphics[width=\linewidth]{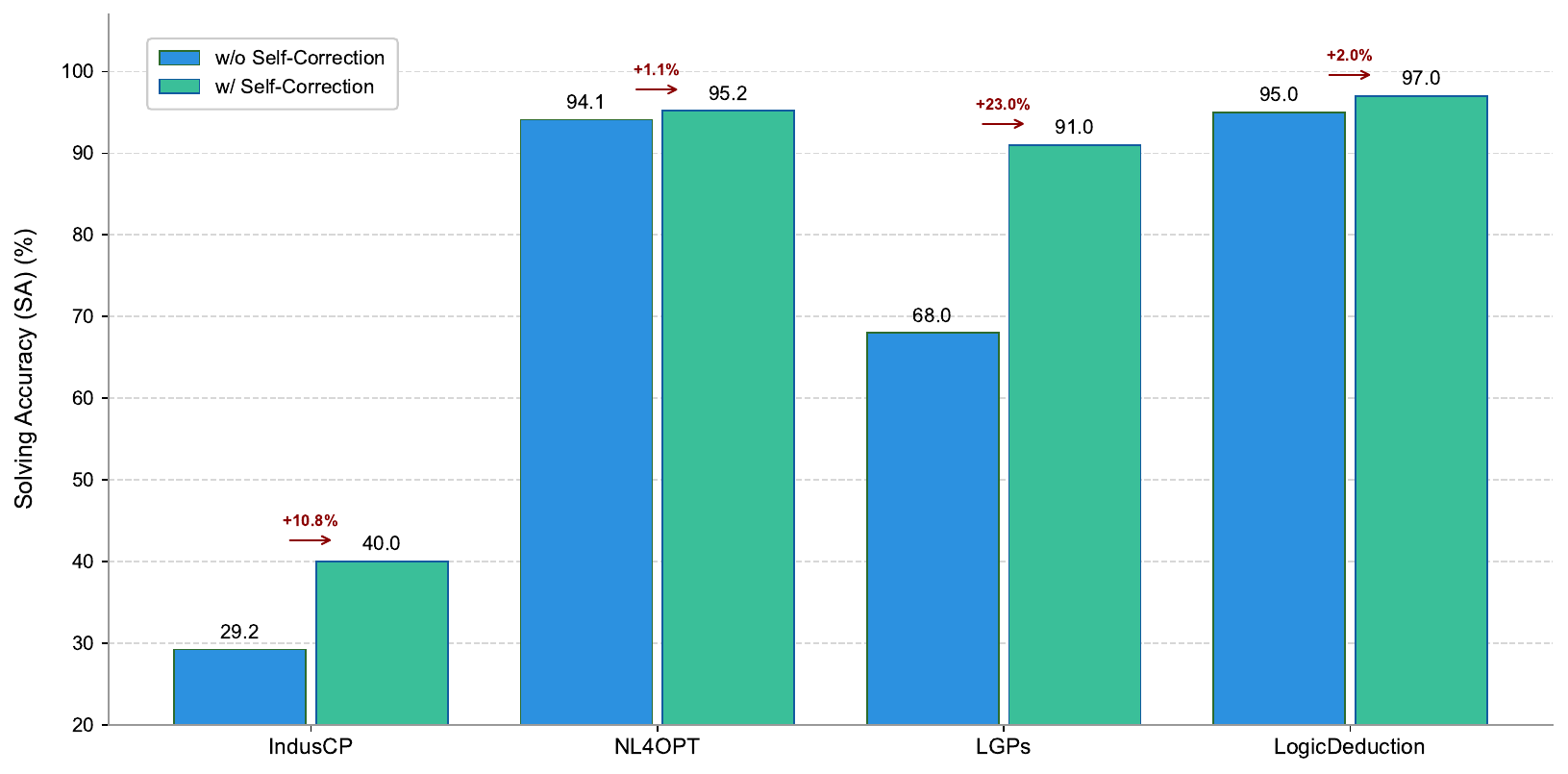}
        \caption{Ablation of Self-Correction}
        \label{fig:sub_image3}
    \end{subfigure}

    \caption{Contribution analysis of key ConstraintLLM components: Comparing the impact of SFT, CARM vs. RAG, and Self-Correction on SA.}
    \label{fig:main_combined_figure_2plus1}
\end{figure*}

\paragraph{CP Solver Implementation Details.}
All models were implemented in Python 3.12 using the PyCSP3~\cite{lecoutre2020pycsp3} library, which interfaces with the Choco Solver~\cite{prud2022choco}. We set a 20-second solver time limit for each test instance.

\paragraph{Training Details.}
Due to limited GPU resources, we \textbf{only use 3 NVIDIA RTX A6000 GPUs} for training and inference, which is highly cost-effective. We utilize Qwen2.5-Coder-32B-Instruct \cite{hui2024qwen2} as our base model to train our Qwen2.5-ConstraintLLM-32B.
Model training is conducted using the framework presented in Llamafactory~\cite{zheng2024llamafactory}, employing QLoRA~\cite{dettmers2023qlora}.
For more details, please see Appendix~\ref{app:sft_detail}.

\paragraph{Baselines.}
We compare ConstraintLLM to three baselines:
(1) \textbf{Direct Solving}: This is the most straightforward baseline. We provide the problem description to the LLM and instruct it to analyze the problem and output the answer.
(2) \textbf{Chain of Thought~(CoT) \cite{wei2022chain}}: To enhance the reasoning abilities of the LLMs, we employ the CoT prompting strategy. The model is prompted to first output a step-by-step thought process before generating the final CP model. Since LLMs can rarely generate successfully executable CP models under \textbf{zero-shot conditions}, we adopted a \textbf{one-shot} setting for the CoT evaluation.
(3) \textbf{Retrieval-Augmented Generation~(RAG) \cite{lewis2020retrieval}}:
This baseline aims to evaluate the utility of a general-purpose retrieval mechanism in aiding LLMs in solving CSPs/COPs. 
Given a target problem description $Q_{\text{NL}}$, we first retrieve $k$ most similar "problem description-model code" pairs from a pre-constructed knowledge base. 
Retrieval is based on the embedding similarity between $Q_{\text{NL}}$ and the problem descriptions in the knowledge base.

\paragraph{Computational Cost.} In terms of practical viability, solving a typical problem from the IndusCP benchmark takes approximately 3 minutes with our framework (w/o ToT) and 7 minutes (w/ ToT). A detailed breakdown of the computational overhead, including LLM inference and solver calls, is provided in Appendix~\ref{sec:appendix_cost_analysis}.

\subsection{RQ1: How does ConstraintLLM Compare to State-of-the-Art Models and Methods}
\label{sec:SOTA}
To comprehensively evaluate the performance of ConstraintLLM, we benchmark it against two distinct categories of State-of-the-Art (SOTA) approaches:

\textbf{1. Comparison with Large-Scale General-Purpose LLMs:}
We first contextualize ConstraintLLM by comparing it against industry-leading, larger-scale general-purpose LLMs, including OpenAI's GPT-4o and Deepseek-V3-685B~\cite{liu2024deepseek}. Our goal was to assess their CP-solving abilities under relatively "out-of-the-box" or minimally prompted conditions. We primarily employed three strategies to elicit CP model code generation: (1) CoT Modeling with one-shot prompting; (2) RAG with four-shot prompting; and (3) our ConstraintLLM framework with four-shot prompting and four rounds of iterative Self-Correction. All comparative experiments were conducted on four datasets: IndusCP, NL4OPT \cite{ramamonjison2023nl4opt}, LGPs \cite{mitra2015learning}, and LogicDeduction \cite{pan2023logic}, using SA as the core evaluation metric. The comprehensive results are presented in Table~\ref{tab:llm_comprehensive_comparison}.

Despite the DeepSeek-V3 LLM having 21.4 times the parameter count of our ConstraintLLM-Qwen2.5-32B model, the latter achieves SOTA performance on three datasets with our ConstraintLLM (w/ ToT) configuration. This demonstrates that ConstraintLLM enables smaller models to exhibit capabilities approaching or even surpassing those of much larger models, significantly bridging the gap in their reasoning abilities.

To provide a more granular view of our model's performance on the IndusCP benchmark, we also analyzed its accuracy across different problem categories. Our model demonstrates stable performance on major categories like scheduling and resource allocation. A detailed breakdown, which also discusses performance on outlier categories with fewer instances or extreme complexity (e.g., Cryptography), is available in Appendix~\ref{sec:appendix_per_category}.

\textbf{2. Comparison with Published SOTA CP Modeling Methods:}
To further position ConstraintLLM within the broader landscape of automated CP modeling and assess its competitiveness, we also benchmarked its performance against several notable SOTA methods published in recent literature. 
This comparison was conducted on the same established public benchmarks relevant to constraint modeling and reasoning. The comparative results, including performance figures reported in the original publications of these SOTA approaches where direct re-implementation was not feasible, are presented in Table~\ref{tab:llm_comprehensive_comparison}.

\subsection{RQ2: Contribution Analysis of Key Components in ConstraintLLM}

To answer RQ2, as illustrated in Figure~\ref{fig:main_combined_figure_2plus1}, we dissect the individual contributions of the core components and strategies within the ConstraintLLM framework to its overall performance through a series of meticulous ablation studies. This section aims to quantify the pivotal roles played by SFT, the CARM, and the iterative Self-Correction mechanism in enhancing the model's ability. 

All ablation experiments were performed on the datasets described in Section~\ref{sec:dataset&metrics}, using SA as the primary evaluation metric. We primarily conducted three sets of comparisons: first, we compared the performance of models with and without our specific SFT within the framework to validate SFT's effectiveness; second, we contrasted generic RAG with our proposed CARM to reveal CARM's advantage in providing high-quality contextual examples; finally, by comparing performance with the self-correction mechanism enabled versus disabled, we quantified its role in fixing errors and improving final solution rates. To further isolate the contribution of SFT, we conducted an additional ablation study comparing our fine-tuned model directly against its base model. The results, detailed in Appendix~\ref{sec:appendix_sft_ablation}.

The experimental results demonstrate the positive impact of each component. Specifically, SFT significantly enhances the model's foundational modeling and understanding capabilities; CARM, compared to generic RAG, retrieves more relevant exemplars, thereby guiding model generation more effectively; and the self-correction mechanism shows strong capabilities in rectifying potential errors in the initial code, all contributing to the superior performance of ConstraintLLM.

\subsection{RQ3: To what extent does ConstraintLLM improve Solving Accuracy in solving CP problems?}
As shown in Table~\ref{tab:llm_comprehensive_comparison}, our ConstraintLLM consistently outperforms all previous methods across multiple challenging benchmarks.
Compared to Chain of Thought (CoT) with one-shot prompting, ConstraintLLM achieves significant improvements in Solving Accuracy (SA).
Specifically, when leveraging ChatGPT-4o as the inference model, ConstraintLLM leads to an average SA improvement of 29.07\%.

Similarly, with DeepSeek-V3 as the inference model, the average SA improvement is 24.02\%.
Furthermore, on our fine-tuned Qwen2.5-Coder-32B model, ConstraintLLM (with ToT) demonstrates an average SA improvement of 38.47\% over its corresponding CoT baseline.
Notably, the most substantial SA improvement rate, reaching 90\%, is observed on the LGPs dataset when applying ConstraintLLM.

\section{Conclusion and Future Work}
\label{sec:conclu}
This work introduces ConstraintLLM, a novel neuro-symbolic framework for industrial-level CP modeling that uses a large language model with multi-instruction SFT, a CARM, and a ToT framework with guided self-correction to improve in-context learning and accuracy. Key contributions are the pioneering training of an open-source LLM for CP modeling and the IndusCP benchmark, where ConstraintLLM achieved SOTA solving accuracy, significantly outperforming baselines. 

Future work will expand ConstraintLLM's capabilities through broader problem domains, refined self-correction, training on larger models, and applying it to more real-world challenges while extending the IndusCP benchmark.


\section*{Limitations}

First, while ConstraintLLM is designed as a general  framework for CP modeling, evaluation is only done on the Python API of PyCSP3 library due to the lack of data in other forms of code.

Second, we trained and evaluated our methods based on the Qwen2.5 series of models. It will be beneficial to extend the baseline model to a broader range of open-source LLMs.

Third, our ConstraintLLM framework relies on CARM (a retrieval-augmented mechanism) to provide in-context exemplars. Ideally, the model should be capable of high-quality CP modeling directly under zero-shot or few-shot conditions, thereby completely obviating the need for an external retrieval database.


\section*{Acknowledgments}
We thank Shenghua Feng, Rui Han, Yu Zhang, and Yuhang Dong for their efforts in dataset annotation. We are also grateful to Hang Gao and the anonymous reviewers for their insightful comments and suggestions, which greatly improved the quality of this paper.

This work was partially supported by NSFC grant No. 62132020.



\bibliography{reference}

\appendix

\section{Background on Symbolic Reasoning Paradigms}
\label{sec:appendix_background}

This appendix provides a brief overview of three fundamental paradigms in symbolic reasoning relevant to our work: Boolean Satisfiability (SAT), Satisfiability Modulo Theories (SMT), and Constraint Programming (CP).

\subsection{Boolean Satisfiability (SAT)}
The Boolean Satisfiability (SAT) problem is foundational, asking whether a satisfying truth assignment exists for a given Boolean formula \cite{biere2009handbook}. Modern solvers, based on Conflict-Driven Clause Learning (CDCL), can efficiently handle instances with millions of variables. A key technique is \textit{bit-blasting}, which translates high-level constraints (e.g., integer arithmetic) into a large SAT formula \cite{jia2023improving}. Maximum Satisfiability (MaxSAT) problem, an optimization variant of SAT, addresses problems requiring the satisfaction of a maximum number of clauses, which has been used in basic tasks of knowledge representation and reasoning \cite{liu2023can,liu2025scalable}.

\subsection{Satisfiability Modulo Theories (SMT)}
Satisfiability Modulo Theories (SMT) extends SAT by integrating specialized "theory solvers" for richer domains like arithmetic, arrays, and bit-vectors \cite{barrett2018satisfiability}. An SMT solver combines a SAT engine for the Boolean structure with theory solvers that check the consistency of constraints within their respective domains. A particularly active area of research is the theory of Nonlinear Real Arithmetic (NRA), crucial for formal verification and synthesis tasks \cite{liu2023nrago,jia2023suggesting,jia2025complete}.

\subsection{Constraint Programming (CP)}
Constraint Programming (CP) is a declarative paradigm distinguished by its highly expressive, high-level modeling language \cite{rossi2006handbook}. Unlike SAT/SMT, which operate on low-level logical formulas, CP allows modelers to use powerful \emph{global constraints} that directly encapsulate complex combinatorial substructures, such as \texttt{AllDifferent}, \texttt{Cumulative}, and \texttt{Circuit}.

This ability to naturally and compactly model real-world problems is the primary reason we focus on CP. While powerful, this high-level modeling process itself presents a significant challenge requiring expertise and precision—the very bottleneck our work is designed to address. CP solvers then use this model, typically employing a combination of search and constraint propagation to efficiently find solutions.
Recent work has also tried to integrate deep learning for solving certain kinds of CP tasks \cite{liu2020learning}.

\section{A Concrete Walkthrough of the ConstraintLLM Pipeline}
\label{sec:appendix_walkthrough}

To illustrate the inner workings of our pipeline, particularly the Constraint-Aware Retrieval Mechanism (CARM) and Guided Self-Correction, we present this step-by-step walkthrough. We use the Traveling Salesman Problem (TSP) as the target problem and the Traveling Purchaser Problem (TPP) as a contrasting example retrieved from the knowledge base.

\subsection{Problem Definitions}

\paragraph{Traveling Salesman Problem (TSP)}
The TSP is a classic optimization problem where the goal is to find the shortest possible tour that visits each city exactly once and returns to the starting city.
\begin{itemize}
    \item \textbf{Input Format}: A symmetric distance matrix, \texttt{distances}, where \texttt{distances[i][j]} is the distance from city \texttt{i} to city \texttt{j}.
\end{itemize}

\paragraph{Traveling Purchaser Problem (TPP)}
A company needs to source a list of materials from suppliers in various cities. The goal is to plan a single circuit trip, starting and ending at the home base, to minimize total expenditure (travel costs + purchase prices). The plan must determine the optimal route and the best city on that route to purchase each material.
\begin{itemize}
    \item \textbf{Input Format}:
    \begin{itemize}
        \item \texttt{nProducts}: The total number of materials.
        \item \texttt{distances}: A matrix of travel costs between cities.
        \item \texttt{prices}: A matrix where \texttt{prices[i][j]} is the cost of material \texttt{j} in city \texttt{i}.
    \end{itemize}
\end{itemize}

\subsection{Pipeline Execution Walkthrough}

Given the natural language description of TSP, the pipeline proceeds as follows:

\begin{enumerate}
    \item \textbf{Constraint Profile Generation:} The model first analyzes the TSP description and generates a "Constraint Profile" containing key constraints it identifies as relevant: \texttt{["Circuit", "Sum", "Element", "Minimum"]}.

    \item \textbf{Constraint-Aware Retrieval (CARM):} Using this profile, CARM retrieves similar problems. For this example, it retrieves four problems, including TPP, which has a similar profile of \texttt{['Circuit', 'Element']}.

    \item \textbf{Guided Self-Correction:} This step demonstrates how the model learns from contrasting examples. Assume the model initially generates an incorrect solution for TSP by misapplying the \texttt{Circuit} constraint.
    
    \begin{enumerate}
        \item \textit{Learning from a Contrasting Example (TPP):} The model analyzes the retrieved "correction path" for TPP. In the TPP case, an initial model using \texttt{AllDifferent} was incorrect. The correction path explains why: \texttt{AllDifferent} would force a visit to \textit{every} city, but TPP only requires visiting a \textit{subset} of cities. The correct TPP model uses \texttt{Circuit}, as it naturally allows for unvisited cities (represented as self-loops where a city variable points to itself, e.g., \texttt{x[i] == i}).
        
        \item \textit{Applying the Insight to TSP:} Armed with this understanding, the model re-evaluates its own TSP model. It recognizes that its use of \texttt{Circuit} is flawed for the opposite reason. For TSP, \texttt{Circuit} is problematic because it can result in multiple disconnected sub-tours, failing to create a single tour of \textit{all} cities. The model deduces that \texttt{AllDifferent} is the correct core constraint for TSP, as it enforces a single permutation of all cities, thus eliminating sub-tours. The correction path then guides the model to replace the flawed \texttt{Circuit} constraint with \texttt{AllDifferent}.
    \end{enumerate}
\end{enumerate}

By understanding why \texttt{Circuit} is right for TPP (allowing subsets) and wrong for TSP (allowing sub-tours), the model learns the underlying modeling principles, not just syntactic patterns.

\section{Ablation Study on Supervised Fine-Tuning (SFT)}
\label{sec:appendix_sft_ablation}

To isolate and directly measure the performance contribution of our Supervised Fine-Tuning process, we conducted an additional ablation study. In this experiment, we evaluated our fine-tuned model without any of the other framework components (i.e., no CARM-based retrieval and no iterative self-correction).

We compared our fine-tuned \texttt{ConstraintLLM-32B} directly against its base model, \texttt{Qwen2.5-Coder-32B}, on the direct modeling task. The models were tasked to generate a CP model from the problem description alone. The results are presented in Table~\ref{tab:sft_ablation_results}.

\begin{table*}[h!]
\centering
\small
\caption{Performance comparison of the fine-tuned model against its base model without other framework components. Results show the direct impact of SFT.}
\label{tab:sft_ablation_results}
\begin{tabular}{lrrrr}
\toprule
\textbf{Model} & \textbf{IndusCP} & \textbf{NL4OPT} & \textbf{LGPs} & \textbf{LogicDeduction} \\
\midrule
Qwen2.5-Coder-32B (Base) & 8.31\% & 79.7\% & 60\% & 79\% \\
ConstraintLLM-32B (SFT) & 12.31\% & 82.55\% & 64\% & 83\% \\
\bottomrule
\end{tabular}
\end{table*}

These results lead to two key conclusions. First, our \texttt{ConstraintLLM-32B} outperforms the base model across all benchmarks, confirming that SFT effectively enhances the model's foundational capabilities. Second, while SFT provides a consistent boost, the standalone performance remains relatively limited, especially on the complex IndusCP benchmark. This finding demonstrates the necessity of the other synergistic components in our framework—namely CARM and iterative self-correction—to achieve robust performance on industrial-level problems.

\section{Training Details}
\label{app:sft_detail}
We performed SFT on the Qwen2.5-Coder-32B-Instruct model using the LLaMA Factory framework \cite{zheng2024llamafactory}, leveraging three NVIDIA A6000 GPUs.
For model and data configuration, we utilized the 
our \texttt{10K} dataset. 
The training was with a cutoff length of 3500 tokens, employing 4 preprocessing workers.
The training spanned 6.0 epochs with a learning rate of 4e-04, a cosine learning rate scheduler, and 500 warmup steps.
The AdamW optimizer was used with a per-device train batch size of 2 and 2 gradient accumulation steps, resulting in a total effective batch size of 12. The maximum gradient norm was capped at 1.0, and gradient checkpointing was enabled.
For fine-tuning, QLoRA was employed with a rank of 32, an alpha of 64, a dropout rate of 0.01, and targeted \texttt{$W_q$} and \texttt{$W_v$} modules. 
BF16 precision was enabled for hardware and precision settings, and 4-bit quantization was performed using the BitsAndBytes method.

\begin{table*}[h!] 
\centering
\caption{Impact of the number of in-context examples (ICL Shots / CARM Top-K) on Solving Accuracy (SA\%) for CoT and ConstraintLLM (w/o ToT, w/o self-correction, with CARM) on the LGPs dataset.}
\label{tab:app_icl_impact}
\small
\begin{tabular}{lccccccc}
\toprule
\multirow{2}{*}{Method} & \multicolumn{7}{c}{Number of In-Context Examples (Shots / CARM Top-K)} \\
\cmidrule(lr){2-8}
& 0-shot & 1-shot & 2-shots & 3-shots & 4-shots & 5-shots \\
\midrule
CoT (Static ICL)       & 0      & 3      & 24      & 33      & 32      & 33      &  \\
ConstraintLLM (CARM Top-K) & 0      & 40     & 64      & 71      & \bestres{89} & 70      &  \\
\bottomrule
\end{tabular}
\end{table*}

\section{Per-Category Performance Analysis on IndusCP}
\label{sec:appendix_per_category}

An aggregated accuracy score can mask significant performance variations across different problem categories. To provide a more transparent and detailed evaluation, we have broken down the performance of our \texttt{ConstraintLLM (w/ ToT)} on the IndusCP benchmark by category. To further contextualize these results, we also report the 'External Knowledge Percentage' for each category in Table~\ref{tab:per_category_performance}, which is the ratio of problem-code pairs in that category to the entire knowledge base.

\begin{table*}[h!]
\centering
\small
\caption{Per-category performance breakdown of \texttt{ConstraintLLM (w/ ToT)} on the IndusCP benchmark, contextualized by the proportion of external knowledge available for each category.}
\label{tab:per_category_performance}
\begin{tabular}{@{}lrrrr@{}}
\toprule
\textbf{Category} & \textbf{Solving Acc. (\%)} & \textbf{\# in IndusCP} & \textbf{IndusCP (\%)} & \textbf{External Knowl. (\%)} \\
\midrule
Scheduling \& Sequencing & 40.26 & 31 & 23.8 & 10.7 \\
Resource Allocation \& Assignment & 53.52 & 23 & 17.7 & 8.1 \\
Combinatorial Puzzles \& Games & 48.15 & 21 & 16.2 & 57.0 \\
Design \& Configuration & 51.02 & 16 & 12.3 & 3.4 \\
Routing \& Logistics & 47.37 & 12 & 9.2 & 6.0 \\
Layout, Packing \& Cutting & 52.38 & 10 & 7.7 & 4.7 \\
Data-Driven Opt. \& Analytics & 40.00 & 6 & 4.6 & 2.0 \\
Cryptography \& Algorithmic Puzzles & 0.00 & 4 & 3.1 & 3.4 \\
Manufacturing \& Production Planning & 83.33 & 4 & 3.1 & 2.7 \\
Telecommunications \& Network Design & 75.00 & 2 & 1.5 & 2.0 \\
Others & 100.00 & 1 & 0.8 & 0.0 \\
\bottomrule
\end{tabular}
\end{table*}

The analysis reveals several key insights. First, our model demonstrates reasonably stable performance across the largest categories. 

Crucially, this detailed breakdown provides strong evidence for the model's ability to generalize beyond simple retrieval, a key feature of our CARM module design. \textbf{Notably, the model achieves its highest accuracy (83.33\%) in "Manufacturing \& Production Planning," a category where the corresponding external knowledge is limited (2.7\%).} Similarly, it performs well above average in "Resource Allocation \& Assignment" (53.52\% accuracy) with only 8.1\% knowledge representation. This demonstrates the model's ability to generalize its reasoning to domains where it has seen very few examples, by matching underlying logical "constraint profiles" rather than surface-level text.

Finally, the poor performance in the \textbf{Cryptography \& Algorithmic Puzzles} category reflects the inherent complexity of these problems. As shown in Table~\ref{tab:crypto_complexity}, instances in this category tend to involve a significantly higher number of constraints and decision variables. Moreover, their problem statements often contain implicit assumptions or require domain-specific knowledge that make them especially challenging for LLMs.

\begin{table*}[h!]
\centering
\small
\caption{Complexity of instances in the Cryptography \& Algorithmic Puzzles category.}
\label{tab:crypto_complexity}
\begin{tabular}{lrr}
\toprule
\textbf{Problem} & \textbf{\# Constraints} & \textbf{\# Decision Variables} \\
\midrule
Cryptanalysis & 21000 & 3300 \\
Speck & 970 & 1100 \\
OptCrypto & 900 & 1650 \\
RotationPuzzle & 400 & 350 \\
\bottomrule
\end{tabular}
\end{table*}

\section{IndusCP Benchmark Construction Details}
\label{sec:appendix_construction_details}

\begin{table*}[t] 
\centering
\small 
\caption{Summary of problem sources for the IndusCP benchmark. We collected problems from well-established academic and competitive programming platforms.}
\label{tab:induscp_sources}
\begin{tabular}{lcl}
\toprule
\textbf{Source Category} & \textbf{\# Problems} & \textbf{Example Problems} \\
\midrule
Minizinc \cite{minizinc-challenge} & 76 & \begin{tabular}[t]{@{}l@{}}ACCAP, BnnPlanner, CELAR, CarpetCutting, Chessboard, CyclicRCPSP...\end{tabular} \\
\addlinespace
XCSP \cite{xcsp-competitions} & 13 & \begin{tabular}[t]{@{}l@{}}AircraftLanding, ClockTriplet, CoinsGrid, Coloring, LargeScaleScheduling...\end{tabular} \\
\addlinespace
PyCSP3-models \cite{pycsp3-models} & 29 & \begin{tabular}[t]{@{}l@{}}Amaze, BinPacking, BoardColoration, Bugs, Cutstock, DakotaFurniture...\end{tabular} \\
\addlinespace
CSPLib \cite{csplib} & 22 & \begin{tabular}[t]{@{}l@{}}Auction, BACP, BusScheduling, CVRP, Diagnosis, GolombRuler, Knapsack...\end{tabular} \\
\bottomrule
\end{tabular}
\end{table*}

The IndusCP benchmark is the result of a rigorous, multi-stage curation process designed to ensure its quality, diversity, and relevance to industrial-level challenges.

\subsection{Stage 1: Initial Collection}
We began by gathering a comprehensive pool of over 340 candidate problems from four reputable and widely recognized sources in the constraint programming community:
\begin{itemize}
    \item \textbf{Minizinc Challenges:} Annual competitions that feature a wide array of complex CP problems.
    \item \textbf{XCSP Competitions:} A standard format and library for combinatorial optimization problems.
    \item \textbf{CSPLib:} A library of test problems for constraint programming.
    \item \textbf{PyCSP3-models:} A collection of CP models written in the Python-based PyCSP3 library.
\end{itemize}

\subsection{Stage 2: Expert-led Curation}
The initial pool of problems was then meticulously curated by domain experts through a multi-step refinement process. This ensured the final benchmark's quality and consistency. The process involved the following steps:
\begin{enumerate}
    \item \textbf{De-duplication and Screening:} We carefully filtered the collected problems to remove duplicates and highly similar variations. This step was crucial for ensuring the diversity of the problems within the benchmark.
    
    \item \textbf{Problem Description Optimization:} We employed a human-in-the-loop process to standardize all problem descriptions. The goal was to create a clear and consistent two-part format for every problem: a general overview followed by a specific input format section. Domain experts first established this standardized style on a subset of problems. Subsequently, we utilized the DeepSeek-V3 model to rewrite the remaining descriptions based on this style. Every machine-generated description underwent a final manual review by experts to verify its quality, clarity, and consistency.
    
    \item \textbf{Test Case Expansion:} To facilitate a robust evaluation of solver capabilities, we developed a comprehensive suite of test cases for each problem. This typically included 2 to 5 distinct instances per problem, covering different scales and edge conditions.
\end{enumerate}

This structured process resulted in the final selection of 140 problems that constitute the IndusCP benchmark.The sources of these problems are summarized in \tablename~\ref{tab:induscp_sources}.


\section{Cross-Domain Knowledge Generalization Study}
\label{sec:appendix_cross_domain}

To rigorously evaluate the cross-domain generalization capability of our framework, particularly the CARM module, we conducted a challenging experiment: solving problems from the NL4OPT and LogicDeduction benchmarks using a knowledge base derived \textbf{exclusively from the IndusCP benchmark}. This setup forces the model to generalize logical structures from complex industrial problems to solve tasks in different domains.

We compared the performance of our ConstraintLLM (w/o ToT) framework—configured with out-of-domain CARM-based retrieval (4 shots) and iterative self-correction (4 rounds)—against two baselines: CoT (one-shot) and RAG (four-shots). To ensure a fair comparison, all methods utilized the same ConstraintLLM-32B model as the underlying inference engine. We also include the performance of our framework using an in-domain knowledge base as a reference. The results are presented in Table~\ref{tab:cross_domain_results}.

\begin{table*}[h!]
\centering
\small
\caption{Cross-domain generalization performance. The knowledge base for all methods was derived exclusively from the IndusCP benchmark.}
\label{tab:cross_domain_results}
\begin{tabular}{lrr}
\toprule
\textbf{Method} & \textbf{NL4OPT (\%)} & \textbf{LogicDeduction (\%)} \\
\midrule
CoT (one-shot) & 85.6 & 83.5 \\
RAG (four-shots) & 91.7 & 17.2 \\
\midrule
\textbf{ConstraintLLM (OUT-OF-DOMAIN)} & \textbf{92.2} & \textbf{94.0} \\
\midrule
ConstraintLLM (IN-DOMAIN) & 95.2 & 96.0 \\
\bottomrule
\end{tabular}
\end{table*}

Even when retrieving knowledge from a completely different domain (IndusCP), our framework significantly outperforms the baselines. Notably, the performance degradation when moving from in-domain to out-of-domain knowledge is minor (a ~3\% drop on NL4OPT and only a 2\% drop on LogicDeduction). This experiment provides evidence that our framework, powered by CARM, is robust and does not simply rely on retrieving superficially similar problems. Instead, it successfully generalizes the underlying logical structures and constraint patterns across diverse domains.

\section{Data Augmentation Details}
\label{app:detail_of_data_augmentation}

Our primary training data comprises a reserved training subset from our IndusCP benchmark.
We augment both problem descriptions and model code. Code augmentation~\cite{yu2022data} involves techniques like variable renaming and equivalent syntactic transformations to generate functionally identical yet diverse code. Problem description augmentation uses EDA \cite{wei2019eda} for semantically consistent textual variations. All augmented code is verified for correctness by solving the original problem instance.
To enhance constraint type extraction for CARM, we construct paired data from training samples in the format: {Natural Language Problem Description, Set of Constraint Types C(Q)}.
Furthermore, for error correction data, we first use a smaller model, Qwen2.5-Coder-7B, to generate initial code for training set problems, collecting diverse erroneous samples (syntactic or logical) via multiple sampling. Subsequently, a larger model, Qwen2.5-coder-32B, synthesizes a detailed "Correction Path" for each pair of erroneous and correct code, detailing the error identification and step-by-step correction process. This dataset, {Problem Description, Incorrect Code, Correction Path, Correct Code}, forms the correction exemplar database $\mathcal{E}$ mentioned in Section~\ref{subsec:Iterative Self-Correction with Guided Retrieval}.
By integrating these diverse data formats and instructions tailored for different objectives (base modeling, constraint extraction, error correction), we perform multi-instruction SFT to comprehensively enhance the model's overall proficiency in constraint solving tasks, the augmented training data contains 10k instances

\section{Generalizability to Other Open-Source Models}
\label{sec:appendix_generalizability}

To assess whether the benefits of our framework extend beyond the fine-tuned Qwen model, we conducted preliminary inference experiments on the IndusCP benchmark using other popular open-source models. Due to resource and time constraints, we did not perform Supervised Fine-Tuning (SFT) on these additional models. 

We evaluated each model under three distinct conditions: (1) a one-shot Chain-of-Thought (CoT) prompt, (2) a standard 4-shot Retrieval-Augmented Generation (RAG) baseline, and (3) our ConstraintLLM (w/o ToT) framework, which uses CARM-based retrieval (4 shots) and iterative self-correction (4 rounds). The results, presented in Table~\ref{tab:generalizability_results}, demonstrate the standalone contribution of our framework's architecture.

\begin{table*}[h!]
\centering
\small
\caption{Results showing the generalizability of the ConstraintLLM framework (w/o ToT) to other open-source models on the IndusCP benchmark.}
\label{tab:generalizability_results}
\begin{tabular}{llr}
\toprule
\textbf{Model} & \textbf{Method} & \textbf{Solving Acc. (\%)} \\
\midrule
\multirow{3}{*}{Mistral-Small-3.2-24B} & CoT one-shot & 9.85 \\& RAG 4-shots & 14.77 \\& \textbf{ConstraintLLM (w/o ToT)} & \textbf{38.46} \\
\midrule
\multirow{3}{*}{Qwen2.5-Coder-7B}    & CoT one-shot & 3.69 \\ & RAG 4-shots & 6.15 \\& \textbf{ConstraintLLM (w/o ToT)} & \textbf{16.31} \\
\midrule
\multirow{3}{*}{Llama-3.1-8B}& CoT one-shot & 1.54 \\& RAG 4-shots & 3.69 \\& \textbf{ConstraintLLM (w/o ToT)} & \textbf{11.08} \\
\midrule
\textbf{ConstraintLLM-32B (Ours)} & \textbf{ConstraintLLM (w/o ToT)} & \textbf{40.00} \\
\bottomrule
\end{tabular}
\end{table*}

\section{Tree of Thought Details}
\label{app:detail_of_tot}
\subsection{Evaluating the Quality of ``Thought Branches''.}
\label{app:evaluating_of_tought_branches}
To effectively guide the search direction and prune unpromising branches, ToT needs to evaluate each generated ``thought branch''. We employ a direct and result-oriented evaluation strategy: the number of passed test cases from a predefined suite.

Specifically, for the model code \( M \) generated by a ``thought branch'', we attempt to compile it and apply it to a standardized set of test cases \( TC = \{tc_1, tc_2, \ldots, tc_N\} \). Each test case represents a concrete instance of the original problem. We count the number of test cases that the model \( M \) successfully solves, which serves as the evaluation score \( V(M) \) for that branch:
\[
V(M) = \sum_{j=1}^{N} \text{Solve}(M, tc_j)
\]
where \(\text{Solve}(M, tc_j)\) is an indicator function:
\[
\text{Solve}(M, tc_j) =
\begin{cases}
1, & \text{if model } M \text{ successfully } \\
    & \text{\parbox{0.7\textwidth}{pass test case $tc_j$ } } \\
0, & \text{otherwise}
\end{cases}
\]
\( V(M) \) directly quantifies the practical effectiveness of the modeling approach represented by a ``thought branch''. A higher number of passed test cases results in a higher score, indicating that the branch is more promising and closer to a correct and efficient final model. This evaluation score is used to guide ToT's selection strategy (e.g., prioritizing branches with higher scores during search) and pruning operations (discarding branches with excessively low scores).
\subsection{Tree of Thought parameter settings}
\label{app:ToT_parameter}
The number of initial thoughts be set to 2. This implies that at the outset of problem-solving, the system will generate two distinct preliminary ideas or approaches. Subsequently, for the expansion at each level of thought, the number of branches to select and further develop from the current level is set to 2, ensuring that the most promising lines of reasoning are pursued in greater depth. Concurrently, the maximum depth of the thought tree is configured to 2, which limits the total number of hierarchical levels in the entire thought process, allowing for an initial set of thoughts followed by at most one level of subsequent reasoning and refinement.

\section{Analysis of OR and CP}
Figure \ref{fig:circular_permutation_problem} presents a comparison of the correct code generated for the same Constraint Optimization Problem when solved using CP and Operations Research (OR), respectively. The OR approach for solving COPs is often complex and error-prone, whereas the CP solution is characteristically concise and elegant.

\begin{figure*}[ht!]
\centering
\fbox{
\begin{minipage}{0.45\linewidth}
\textbf{IndusCP}: We have a circular arrangement of 12 distinct numbers (from 1 to 12) that need to be rearranged in a specific order. The goal is to ensure that the sum of any three consecutive numbers in the circle is as small as possible. The arrangement must meet the following conditions: the number 1 must be fixed in the first position, the second number must be smaller than the last number (to break symmetry), and every number from 1 to 12 must appear exactly once.
\end{minipage}
}
\hfill
\fbox{
\begin{minipage}{0.45\linewidth}
\textbf{NL4OPT}: A fishery wants to transport their catch. They can either use local sled dogs or trucks. Local sled dogs can take 100 fish per trip while trucks can take 300 fish per trip. The cost per trip for sled dogs is \$50 while the cost per trip for a truck is \$100. The budget is at most \$1000 and the number of sled dog trips must be less than the number of truck trips. Formulate an LP to maximize the number of fish that can be transported.
\end{minipage}
}

\vspace{0.8em}

\fbox{
\begin{minipage}{0.45\linewidth}
\textbf{LGPs}: 
Clues: 
\begin{itemize}
\item Vicky Estes used the catamaran.
\item Debra Decker took 4 fewer days to finish than the sailor in the trimaran.
\item Wendell Orr finished in 278 days.
\end{itemize}
Entities: 
\begin{itemize}
\item Days: 270, 274, 278, 282
\item Boat types: catamaran, ketch, schooner, trimaran
\item Sailors: Debra Decker, Gil Baxter, Vicky Estes, Wendell Orr
\end{itemize}
\end{minipage}
}
\hfill
\fbox{
\begin{minipage}{0.45\linewidth}
\textbf{LogicalDeduction}: On a shelf, there are five books: a green book, a blue book, a white book, a purple book, and a yellow book. The blue book is to the right of the yellow book. The white book is to the left of the yellow book. The blue book is the second from the right. The purple book is the second from the left. Which of the following is true?
\begin{itemize}
\item[A)] The green book is the second from the left.
\item[B)] The blue book is the second from the left.
\item[C)] The white book is the second from the left.
\end{itemize}
\end{minipage}
}
\caption{Four representative benchmark examples from our evaluation, showing the diversity of problem types.}
\label{fig:benchmark_examples}
\end{figure*}

\section{Experimental Details}
We use OpenAI's \texttt{text-embedding-ada-002} model to generate text embeddings. Furthermore, we maintain the same prompt and other relevant hyperparameters  as used in ConstraintLLM.
\subsection{Dataset Details}
\label{app:datasets_detail}

\textbf{IndusCP}: As detailed in Section~\ref{sec:indus_cp_benchmark}, this is our constructed industrial-level benchmark for constraint satisfaction problems comprises 140 curated problem instances, each instance includes 2 to 5 distinct test cases to validate the correctness of modeling.

\textbf{NL4OPT} \cite{ramamonjison2023nl4opt}: This dataset provides natural language descriptions of linear optimization problems. While its focus on linear programming differs from the declarative and combinatorial nature of CSPs, it serves as a relevant reference for evaluating an LLM's understanding of structured optimization problems from text. We select 271 instances for the test set and 713 instances for the training set.

\textbf{LGPs} \cite{mitra2015learning}: This dataset consists of logical puzzles described with clues and entities, which can be formulated as CSPs. We select 50 instances as train data and 100 instance as test data.

\textbf{LogicDeduction} \cite{pan2023logic}: A complex logical inference task from the BigBench benchmark suite \cite{srivastava2022beyond}, can be expressed as CSPs. The core challenge involves determining the correct sequence of objects given a limited set of premises. 
For our experiments, we used the test set of 200 instances.

\section{Baseline Implementation Details}
\label{sec:appendix_baselines}

This section provides the detailed implementation specifics for the Chain-of-Thought (CoT) and Retrieval-Augmented Generation (RAG) baselines used in our experiments. The full prompts used are detailed in Figure~\ref{fig:cot_prompt} and Figure~\ref{fig:rag_prompt}.

\lstset{
    basicstyle=\ttfamily\small, 
    breaklines=true,            
    breakatwhitespace=false,    
    frame=single,               
    showstringspaces=false,     
    numbers=none,               
    xleftmargin=1em,            
    xrightmargin=1em            
}

\begin{figure*}[t!]
\begin{lstlisting}
You are a Python programming expert capable of using the pycsp3 library to solve Constraint Satisfaction Problems (CSPs). Based on the given input data, generate a syntactically and semantically correct constraint solving model. 
Instructions:
- think step by step,use chain of thought
- Based on data input: Your generated code must assume that a variable named data exists, which contains the input data required for the problem. The first step of your code must be to unpack data into meaningful variables. For example: n, edges, start, end = data or marioHouse, luigiHouse, fuelLimit, houses = data
- Extraction of necessary inputs from the data
- Use only the pycsp3 library: `from pycsp3 import *`
- code should use ```python your_code_here``` to wrap the code

a example:
Problem Description:
The problem is to assign a set of n nodes to m available rings. Each ring has a maximum capacity, r, which is the highest number of nodes it can accommodate. A specific subset of these nodes, identified in a connections list, must be assigned to exactly one ring. The overall objective is to find a valid assignment that respects the capacity of each ring and the single-assignment rule for the specified connection nodes, while minimizing the total number of node-to-ring assignments. To ensure a unique solution and avoid symmetrical configurations, the assignments across the rings must follow a lexicographically increasing order.

Input Format:
n: An integer representing the total number of nodes in the network.
m: An integer representing the number of available rings for assignment.
r: An integer representing the maximum capacity for each ring.
connections: A list of integers specifying the nodes that must be assigned to exactly one ring.

code:
from pycsp3 import *
n, m, r, connections = data

x = VarArray(size=[m, n], dom={0, 1})
T = {tuple(1 if j // 2 == i else ANY for j in range(2 * m)) for i in range(m)}

satisfy(
    [(x[i][conn] for i in range(m)) in T for conn in connections],
    [Sum(x[i]) <= r for i in range(m)],
    LexIncreasing(x)
)
minimize(
    Sum(x)
)
\end{lstlisting}
\caption{The full one-shot prompt used for the Chain-of-Thought (CoT) baseline.}
\label{fig:cot_prompt}
\end{figure*}

\begin{figure*}[t!]
\begin{lstlisting}
You are a Python programming expert capable of using the pycsp3 library to solve Constraint Satisfaction Problems (CSPs). Based on the given input data, generate a syntactically and semantically correct constraint solving model. 
Instructions:
- Based on data input: Your generated code must assume that a variable named data exists, which contains the input data required for the problem. The first step of your code must be to unpack data into meaningful variables. For example: n, edges, start, end = data or marioHouse, luigiHouse, fuelLimit, houses = data
- Extraction of necessary inputs from the data
- Definition of the objective function (minimize/maximize)
- Use only the pycsp3 library: `from pycsp3 import *`
- code should use ```python your_code_here``` to wrap the code

Here are some similar questions and codes for reference:
### Example 1
Problem: [Retrieved Problem Description 1]
Code: [Retrieved Code 1]

### Example 2
... (up to Example 4)

The Problem: 
[Current Test Problem Description]
\end{lstlisting}
\caption{The prompt template used for the Retrieval-Augmented Generation (RAG) baseline.}
\label{fig:rag_prompt}
\end{figure*}

\subsection{Retrieval-Augmented Generation (RAG) Baseline}
Our RAG implementation is a standard two-stage process constructed using the LangChain framework.

\paragraph{Stage 1: Retrieval.}
For the retrieval stage, the \textbf{knowledge base} for each benchmark was composed of `(Problem Description, Correct CP Model Code)` pairs. Problem descriptions were split into documents (700-token \texttt{chunk\_size}, 100-token \texttt{chunk\_overlap}) and vectorized using OpenAI's \texttt{text-embedding-ada-002} \textbf{embedding model}. These embeddings were indexed in a FAISS \textbf{vector store}. We used \textbf{Cosine Similarity} as the metric to retrieve the \textbf{top-4} most relevant pairs for each query.

\paragraph{Stage 2: Generation (Prompting).}
The four retrieved examples were formatted into the few-shot prompt template shown in Figure~\ref{fig:rag_prompt}.

\section{Number of In-Context Examples}
\label{app:number_icl}
In this appendix, we further investigate the impact of varying numbers of In-Context Learning examples on model performance in CP modeling tasks. We primarily focus on two scenarios: (1) providing a varying number of static ICL examples under standard Chain-of-Thought prompting; and (2) the effect of the number of Top-K examples retrieved by CARM within our ConstraintLLM framework (without ToT and self-correction, but with CARM enabled for dynamic retrieval). All experiments were conducted on the LGPs dataset, using Solving Accuracy (SA\%) as the evaluation metric.

For the CoT experiments, we directly included 0 to 5 static, task-relevant ICL examples in the prompt. For the ConstraintLLM (w/o ToT, w/o self-correction, but with CARM) experiments, the number of "shots" corresponds to the Top-K most relevant examples retrieved by CARM and provided to the LLM. In both cases, 0-shot indicates that the model relies on no explicitly provided contextual examples for inference.

Table~\ref{tab:app_icl_impact} presents the impact of different numbers of ICL examples on the performance of CoT and ConstraintLLM (CARM Top-K) on the LGPs dataset. 
The results indicate that under zero-shot conditions, the LLM is unable to generate CP models that successfully solve the problems, thereby demonstrating the necessity of using In-Context Learning.

\section{Computational Cost Analysis}
\label{sec:appendix_cost_analysis}

Understanding the computational cost is essential to assess the practical viability of our framework. We analyzed the costs based on our experiments on the 140 problems in the IndusCP benchmark. Our analysis was performed on the ConstraintLLM (w/o ToT) framework, configured with a maximum of 4 self-correction rounds. The primary cost drivers are LLM inference and solver calls, as detailed in Table~\ref{tab:computational_cost}.

\begin{table*}[t!]
\centering
\small
\caption{Computational cost breakdown per problem on the IndusCP benchmark for ConstraintLLM (w/o ToT).}
\label{tab:computational_cost}
\begin{tabular}{@{}lrrrr@{}}
\toprule
\textbf{Component} & \textbf{\begin{tabular}[c]{@{}c@{}}Avg. \# Calls \\ per Problem\end{tabular}} & \textbf{\begin{tabular}[c]{@{}c@{}}Avg. Time \\ per Call (s)\end{tabular}} & \textbf{\begin{tabular}[c]{@{}c@{}}Max \# Calls \\ per Problem\end{tabular}} & \textbf{\begin{tabular}[c]{@{}c@{}}Max Time \\ per Call (s)\end{tabular}} \\
\midrule
\textbf{LLM Inference}        & 5.0 (700/140)  & 29.55    & 6 & 42.25 \\
\textbf{Solver}               & 2.9 (411/140)  & 1.54     & 5 & 20.025 \\
\textbf{CARM Retrievals}      & 2.5 (349/140)  & 0.000173 & 5 & 0.000564 \\
\textbf{Self-Correction Loops} & 1.4 rounds (196/140) & 41.6 & 4 rounds & 89.74 \\
\bottomrule
\end{tabular}
\end{table*}

\paragraph{Overall Cost and Upper Bound for ConstraintLLM (w/o ToT).}
The theoretical upper bound for the time complexity is given by:
\begin{equation}
\label{eq:cost_no_tot}
\resizebox{0.9\columnwidth}{!}{$
    T_{\text{upper}}(P) = O\Big(k \cdot L_{\text{gen}}(P) \cdot f(t_{\text{token}}) + k \cdot T_{\text{solver}}(P) + k \cdot T_{\text{CARM}} \Big)
$}
\end{equation}
Where:
\begin{itemize}
    \item $P$: A specific problem instance.
    \item $k$: The maximum number of self-correction iterations.
    \item $L_{\text{gen}}(P)$: The maximum number of tokens generated in any single LLM call for problem $P$.
    \item $T_{\text{solver}}(P)$: The time for a single solver invocation for problem $P$, subject to a timeout.
    \item $T_{\text{CARM}}$: The cost of a single CARM retrieval (typically negligible).
    \item $f(t_{\text{token}})$: The average time to generate a single LLM token.
\end{itemize}
On the IndusCP benchmark, solving a problem with ConstraintLLM (w/o ToT) takes approximately \textbf{3 minutes} on average.

\paragraph{Overall Cost and Upper Bound for ConstraintLLM (w/ ToT).}
The upper bound is significantly higher due to the tree search:
\begin{equation}
\label{eq:cost_with_tot}
\resizebox{0.9\columnwidth}{!}{$
    T_{\text{upper}}(P) = O\Big( N_{\text{nodes}} \cdot (k \cdot L_{\text{gen}}(P) \cdot f(t_{\text{token}}) + k \cdot T_{\text{solver}}(P) + k \cdot T_{\text{CARM}}) \Big)
$}
\end{equation}
Where $N_{\text{nodes}}$ is the total number of nodes explored in the Tree of Thoughts, given by $W \cdot (m^n - 1) / (m - 1)$ for a tree with initial width $W$, branching factor $m$, and depth $n$.
On the IndusCP benchmark, solving a problem with ConstraintLLM (w/ ToT) takes approximately \textbf{7 minutes} on average.

\section{Examples of benchmarks}
Figure~\ref{fig:benchmark_examples} shows four representative benchmark examples from our evaluation.

\lstset{
  basicstyle=\ttfamily\small,
  keywordstyle=\color{blue},
  commentstyle=\color{green},
  stringstyle=\color{red},
  showstringspaces=false
}

\begin{figure*}[ht!]
\centering
\begin{minipage}{0.85\textwidth}
    \begin{tcolorbox}[colframe=blue!50!black, colback=blue!10!white, title=Problem Description]
        We have a circular arrangement of 12 distinct numbers (from 1 to 12) that need to be rearranged in a specific order. The goal is to ensure that the sum of any three consecutive numbers in the circle is as small as possible. The arrangement must meet the following conditions: the number 1 must be fixed in the first position, the second number must be smaller than the last number (to break symmetry), and every number from 1 to 12 must appear exactly once. The task is to find the optimal order that minimizes the largest sum among all possible groups of three consecutive numbers in the circle.
    \end{tcolorbox}
\end{minipage}

\begin{minipage}{0.85\textwidth}
    \begin{tcolorbox}[colframe=red!50!black, colback=red!10!white, title=OR model]
        \begin{lstlisting} % 使用上面 \lstset 定义的全局设置

...
def assign_value_to_position_rule(model, i):
    return sum(model.y[i, v] for v in model.V_set) == 1
model.assign_value_con = pyo.Constraint(model.N_set, rule=assign_value_to_position_rule)
def use_value_in_position_rule(model, v):
    return sum(model.y[i, v] for i in model.N_set) == 1
model.use_value_con = pyo.Constraint(model.V_set, rule=use_value_in_position_rule)
def get_x_value_at_pos(model, pos_idx):
    return sum(val * model.y[pos_idx, val] for val in model.V_set)
model.subsequence_sum_con = pyo.ConstraintList()
for i in model.N_set:
    current_subsequence_sum = sum(get_x_value_at_pos(model, (i + k) % n_val) for k in range(r_val))
model.subsequence_sum_con.add(
current_subsequence_sum <= model.z)
model.y[0, 1].fix(1)
if n_val > 2:
    val_at_pos_1 = get_x_value_at_pos(model, 1)
    val_at_pos_n_minus_1 = get_x_value_at_pos(model, n_val - 1)
    model.symmetry_break_order_con = pyo.Constraint(expr=val_at_pos_1 <= val_at_pos_n_minus_1 - 1)
model.objective = pyo.Objective(expr=model.z, sense=pyo.minimize)
...
        \end{lstlisting}
    \end{tcolorbox}
\end{minipage}

\begin{minipage}{0.85\textwidth}
    \begin{tcolorbox}[colframe=green!50!black, colback=green!10!white, title=CP model]
       
        \begin{lstlisting} % 使用上面 \lstset 定义的全局设置

from pycsp3 import *
r, n = data
x = VarArray(size=n, dom=range(1, n + 1))
z = Var(dom=range(1, n * r + 1))
satisfy(
AllDifferent(x),
x[0] == 1,x[1] < x[-1],
[Sum(x[(i + k) % n] for k in range(r)) <= z for i in range(n)]
)
minimize(z)
        \end{lstlisting}
    \end{tcolorbox}
\end{minipage}

\caption{A comparative analysis of OR and CP models in terms of their code implementation.}
\label{fig:circular_permutation_problem}
\end{figure*}

\end{document}